\documentclass{article}
\usepackage{amsmath, amsthm, amssymb}

\usepackage{expl3}  %
\ExplSyntaxOn
  _to_str:N
\ExplSyntaxOff
\usepackage{ifthen}
\usepackage{pgffor}

\usepackage[acronym,nowarn,section,nonumberlist]{glossaries}
\setacronymstyle{long-sc-short}
\glsdisablehyper  %
\makeglossaries

\setacronymstyle{long-short}

\newacronym{auc}{AUC}{Area Under the Curve}
\newacronym{asr}{ASR}{Automatic Speech Recognition}

\newacronym{bbb}{BBB}{Bayes by Backprop}
\newacronym{bert}{BERT}{Bidirectional Encoder Representations from Transformers}
\newacronym{bow}{BOW}{Bag of Words}
\newacronym{boe}{BOE}{Bag of Embeddings}
\newacronym{borep}{BOREP}{Bag of Random Embedding Projections}
\newacronym{bn}{BN}{Bayesian Network}
\newacronym{byol}{BYOL}{Bootstrap Your Own Latent}
\newacronym{byov}{BYOV}{Bootstrap Your Own Variance}
\newacronym{snr}{SNR}{signal to noise ratio}

\newacronym{cdf}{CDF}{Cumulative Distribution Function}
\newacronym{cnn}{CNN}{Convolutional Neural Network}
\newacronym{cka}{CKA}{Centered Kernel Alignment}
\newacronym{ctc}{CTC}{Connectionist Temporal Classification}
\newacronym{cvae}{CVAE}{Conditional Variational Auto-Encoder}
\newacronym{cv}{CV}{Computer Vision}

\newacronym{dag}{DAG}{Directed Acyclic Graph}
\newacronym{ddpg}{DDPG}{Deep Deterministic Policy Gradient}
\newacronym{dino}{DINO}{DIstillation with NO labels}
\newacronym{dgm}{DGM}{Directed Graphical Model}
\newacronym{dnn}{DNN}{Deep Neural Network}

\newacronym{ece}{ECE}{Expected Calibration Error}
\newacronym{ecfp4}{ECFP4}{Extended Connectivity Fingerprint}
\newacronym{ehr}{EHR}{Electronic Health Record}
\newacronym{elbo}{ELBO}{Evidence Lower Bound}
\newacronym{ema}{EMA}{Exponential Moving Average}
\newacronym{esn}{ESN}{Echo State Network}
\newacronym{esr}{ESR}{Exponential Scaling Rule}

\newacronym{fft}{FFT}{Fast Fourier Transform}
\newacronym{fs}{FS}{FastSent}
\newacronym{ft}{FT}{Fourier Transform}

\newacronym{gat}{GAT}{Graph Attention Network}
\newacronym{gcn}{GCN}{Graph Convolutional Network}
\newacronym{gd}{GD}{Gradient Descent}
\newacronym{gdl}{GDL}{Geometric Deep Learning}
\newacronym{ggnn}{GGNN}{Gated Graph Neural Network}
\newacronym{glove}{GloVe}{Global Vectors for Word Representation}
\newacronym{gnn}{GNN}{Graph Neural Network}
\newacronym{gru}{GRU}{Gated Recurrent Unit}

\newacronym{hmm}{HMM}{Hidden Markov Model}

\newacronym{idf}{IDF}{Inverse Document Frequency}
\newacronym{ig}{IG}{Information Gain}
\newacronym{iid}{i.i.d.}{Independent and Identically Distributed}
\newacronym{isan}{ISAN}{Input Switched Affine Network}

\newacronym{jsd}{JSD}{Jensen-Shannon Divergence}

\newacronym{kld}{KLD}{Kullback-Leibler Divergence}
\newacronym{knn}{KNN}{$K$-Nearest Neighbour}

\newacronym{lhs}{L.H.S.}{left hand side}
\newacronym{lstm}{LSTM}{Long Short Term Memory Network}
\newacronym{lsr}{LSR}{Linear Scaling Rule}

\newacronym{map}{MAP}{Maximum a Posteriori}
\newacronym{mc}{MC}{Monte Carlo}
\newacronym{mcmc}{MCMC}{Markov chain Monte Carlo}
\newacronym{mdl}{MDL}{Minimum Description Length}
\newacronym{ml}{ML}{Machine Learning}
\newacronym{mlp}{MLP}{Multi-Layer Perceptron}
\newacronym{mle}{MLE}{Maximum Likelihood Estimation}
\newacronym{mnist}{MNIST}{Modified National Institute of Standards and Technology}

\newacronym{nlp}{NLP}{Natural Language Processing}
\newacronym{nn}{NN}{Neural Network}
\newacronym{nll}{NLL}{Negative Log Likelihood}

\newacronym{ode}{ODE}{Ordinary Differential Equation}
\newacronym{oh}{OH}{One-Hot}
\newacronym{ood}{OOD}{out-of-distribution}

\newacronym{pdf}{PDF}{Probability Density Function}
\newacronym{pgm}{PGM}{Probabilistic Graphical Model}
\newacronym{pl}{PL}{Pseudo-Label}

\newacronym{relu}{ReLU}{Rectified Linear Unit}
\newacronym{rdf}{RDF}{Resource Description Framework}
\newacronym{rgb}{RGB}{Red Green Blue}
\newacronym{rgc}{RGC}{Relational Graph Convolution}
\newacronym{rgat}{RGAT}{Relational Graph Attention Network}
\newacronym{rgcn}{RGCN}{Relational Graph Convolutional Network}
\newacronym{rhs}{R.H.S.}{Right Hand Side}
\newacronym{rl}{RL}{Reinforcement Learning}
\newacronym{rnn}{RNN}{Recurrent Neural Network}
\newacronym{roc}{ROC}{Receiver Operating Characteristic}

\newacronym{sde}{SDE}{Stochastic Differential Equation}
\newacronym{sgd}{SGD}{Stochastic Gradient Descent}
\newacronym{sg}{SG}{SkipGram}
\newacronym{simclr}{SimCLR}{Simple framework for Contrastive Learning of visual Representations}
\newacronym{ssl}{SSL}{Self-Supervised Learning}
\newacronym{st}{ST}{SkipThought}
\newacronym{sts}{STS}{Semantic Textual Similarity}
\newacronym{svi}{SVI}{Stochastic Variational Inference}
\newacronym{swa}{SWA}{Stochastic Weight Averaging}
\newacronym{swag}{SWAG}{SWA-Gaussian}

\newacronym{termfreq}{TF}{Term Frequency}
\newacronym{tfidf}{TF-IDF}{Term Frequency-Inverse Document Frequency}
\newacronym[plural=TPPs, descriptionplural={Temporal Point Processes}]{tpp}{TPP}{Temporal Point Process}

\newacronym{vae}{VAE}{Variational Auto-Encoder}
\newacronym{vi}{VI}{Variational Inference}
\newacronym{vit}{ViT}{Vision Transformer}

\newacronym{wer}{WER}{Word Error Rate}
\newacronym{wirgat}{WIRGAT}{Within-Relation Graph Attention}
\newacronym{wl}{WL}{Weisfeiler-Lehman}

\usepackage{bm}
\usepackage[T1]{fontenc}  %
\usepackage[type1]{libertine}
\usepackage[libertine]{newtxmath}
\usepackage{mathtools}

\usepackage[hidelinks]{hyperref}
\usepackage{cleveref}
\usepackage{caption}
\usepackage{subcaption}
\usepackage{booktabs}       %

\usepackage{sidecap}

\usepackage[font=small]{caption}
\usepackage{xcolor}

\usepackage{enumitem}

\usepackage{placeins}
\usepackage{titletoc}
\usepackage[page,header]{appendix}

\usepackage{algorithm}
\usepackage{algpseudocode}

\usepackage{ifthen}

\usepackage{tikz}
\usetikzlibrary{positioning,decorations.pathmorphing,decorations.pathreplacing, fit}

\usepackage{xcolor}
\definecolor{AppleWhite}{RGB}{255,255,255}
\definecolor{ApplePrimaryCoolGrey}{RGB}{116,128,139}
\definecolor{AppleCoolGray1}{RGB}{199,209,214}
\definecolor{AppleCoolGray2}{RGB}{147,174,190}
\definecolor{AppleCoolGray3}{RGB}{124,147,160}
\definecolor{AppleCoolGray4}{RGB}{92,102,109}
\definecolor{AppleCoolGray5}{RGB}{78,93,100}
\definecolor{AppleCoolGray6}{RGB}{53,60,65}
\definecolor{AppleBlack}{RGB}{0,0,0}
\definecolor{AppleSecondaryChartGray}{RGB}{168,168,168}
\definecolor{AppleChartGrey2}{RGB}{233,233,233}
\definecolor{AppleChartGrey3}{RGB}{211,211,211}
\definecolor{AppleChartGrey4}{RGB}{190,190,190}
\definecolor{AppleChartGrey5}{RGB}{140,140,140}
\definecolor{AppleChartGrey6}{RGB}{102,102,102}
\definecolor{AppleChartGrey7}{RGB}{64,64,64}
\definecolor{ApplePrimaryChartBlue}{RGB}{84,151,193}
\definecolor{AppleBlue2}{RGB}{212,229,239}
\definecolor{AppleBlue3}{RGB}{169,202,223}
\definecolor{AppleBlue4}{RGB}{127,177,209}
\definecolor{AppleBlue5}{RGB}{71,130,166}
\definecolor{AppleBlue6}{RGB}{55,99,128}
\definecolor{AppleBlue7}{RGB}{45,72,89}
\definecolor{ApplePrimaryChartGreen}{RGB}{83,172,121}
\definecolor{AppleGreen2}{RGB}{212,234,221}
\definecolor{AppleGreen3}{RGB}{169,213,188}
\definecolor{AppleGreen4}{RGB}{126,193,155}
\definecolor{AppleGreen5}{RGB}{58,140,82}
\definecolor{AppleGreen6}{RGB}{39,102,54}
\definecolor{AppleGreen7}{RGB}{29,58,31}
\definecolor{ApplePrimaryChartYellow}{RGB}{253,195,93}
\definecolor{AppleYellow2}{RGB}{254,240,214}
\definecolor{AppleYellow3}{RGB}{254,224,174}
\definecolor{AppleYellow4}{RGB}{254,210,134}
\definecolor{AppleYellow5}{RGB}{230,168,69}
\definecolor{AppleYellow6}{RGB}{191,131,46}
\definecolor{AppleYellow7}{RGB}{153,107,54}
\definecolor{ApplePrimaryChartOrange}{RGB}{250,151,92}
\definecolor{AppleOrange2}{RGB}{254,229,214}
\definecolor{AppleOrange3}{RGB}{252,203,173}
\definecolor{AppleOrange4}{RGB}{252,178,133}
\definecolor{AppleOrange5}{RGB}{227,121,68}
\definecolor{AppleOrange6}{RGB}{191,87,46}
\definecolor{AppleOrange7}{RGB}{143,59,36}
\definecolor{ApplePrimaryChartRed}{RGB}{227,94,105}
\definecolor{AppleRed2}{RGB}{248,215,217}
\definecolor{AppleRed3}{RGB}{241,174,180}
\definecolor{AppleRed4}{RGB}{234,135,143}
\definecolor{AppleRed5}{RGB}{196,63,77}
\definecolor{AppleRed6}{RGB}{153,35,53}
\definecolor{AppleRed7}{RGB}{102,19,43}
\definecolor{ApplePrimaryChartPurple}{RGB}{161,150,204}
\definecolor{ApplePurple2}{RGB}{231,228,242}
\definecolor{ApplePurple3}{RGB}{208,202,229}
\definecolor{ApplePurple4}{RGB}{185,176,217}
\definecolor{ApplePurple5}{RGB}{128,113,171}
\definecolor{ApplePurple6}{RGB}{89,76,128}
\definecolor{ApplePurple7}{RGB}{62,46,101}
\definecolor{AppleCoolGrey}{RGB}{116,128,139}
\definecolor{AppleChartGray}{RGB}{168,168,168}
\definecolor{AppleBlue}{RGB}{84,151,193}
\definecolor{AppleGreen}{RGB}{83,172,121}
\definecolor{AppleYellow}{RGB}{253,195,93}
\definecolor{AppleOrange}{RGB}{250,151,92}
\definecolor{AppleRed}{RGB}{227,94,105}
\definecolor{ApplePurple}{RGB}{161,150,204}

\DeclareRobustCommand{\mathup}[1]{\begingroup\changegreek\mathrm{#1}\endgroup}
\DeclareRobustCommand{\mathbfup}[1]{\begingroup\changegreekbf\mathbf{#1}\endgroup}
\DeclareRobustCommand{\mathbit}[1]{\bm{\mathit{#1}}}

\DeclareMathAlphabet{\mathsfit}{\encodingdefault}{\sfdefault}{m}{sl}
\SetMathAlphabet{\mathsfit}{bold}{\encodingdefault}{\sfdefault}{bx}{n}
\newcommand{\tens}[1]{\bm{\mathsfit{#1}}}

\newcommand{\constantvector}{\bm}               %

\newcommand{\constantmatrix}{\bm}               %
\newcommand{\constantmatrixgreek}{\mathbit}

\newcommand{\randomscalar}{\textnormal}         %
\newcommand{\randomscalargreek}{\mathup}

\newcommand{\randomvector}{\mathbf}             %
\newcommand{\randomvectorgreek}{\mathbfup}

\newcommand{\randommatrix}{\mathbf}             %
\newcommand{\randommatrixgreek}{\mathbfup}

\newcommand{\graphstyle}{\mathcal}              %

\newcommand{\tensorstyle}{\tens}                %

\newcommand{\setstyle}{\mathbb}                %

\def\alphabet{a,b,c,d,e,f,g,h,i,j,k,l,m,n,o,p,q,r,s,t,u,v,w,x,y,z}
\def\Alphabet{A,B,C,D,E,F,G,H,I,J,K,L,M,M,O,P,Q,R,S,T,U,V,W,X,Y,Z}
\def\greekalphabet{alpha,beta,gamma,delta,epsilon,varepsilon,zeta,eta,theta,vartheta,iota,kappa,varkappa,lambda,mu,nu,xi,pi,varpi,rho,varrho,sigma,varsigma,tau,upsilon,phi,varphi,chi,psi,omega}
\def\GreekAlphabet{Gamma,Delta,Theta,Lambda,Xi,Pi,Sigma,Upsilon,Phi,Psi,Omega}

\makeatletter
\def\changegreek{\@for\next:=\greekalphabet
	\do{\expandafter\let\csname\next\expandafter\endcsname\csname\next up\endcsname}}
\def\changegreekbf{\@for\next:=\greekalphabet
	\do{\expandafter\def\csname\next\expandafter\endcsname\expandafter{%
			\expandafter\bm\expandafter{\csname\next up\endcsname}}}}
\makeatother

\foreach \x in \alphabet {
	\expandafter\xdef\csname v\x\endcsname{\noexpand\ensuremath{\noexpand\constantvector{\x}}}

	\expandafter\xdef\csname ev\x\endcsname{\noexpand\ensuremath{\noexpand\x}}

	\ifthenelse{\equal{\x}{m}}{}{
		\expandafter\xdef\csname r\x\endcsname{\noexpand\ensuremath{\noexpand\randomscalar{\x}}}
	}

	\expandafter\xdef\csname rv\x\endcsname{\noexpand\ensuremath{\noexpand\randomvector{\x}}}
}

\foreach \x in \greekalphabet {
	\expandafter\xdef\csname v\x\endcsname{\noexpand\ensuremath{\noexpand\constantvector{\csname \x\endcsname}}}

	\expandafter\xdef\csname ev\x\endcsname{\noexpand\ensuremath{\noexpand{\csname \x \endcsname}}}

	\expandafter\xdef\csname r\x\endcsname{\noexpand\ensuremath{\noexpand\randomscalargreek{\csname \x\endcsname}}}

	\expandafter\xdef\csname rv\x\endcsname{\noexpand\ensuremath{\noexpand\randomvectorgreek{\csname \x\endcsname}}}
}

\foreach \x in \Alphabet {
	\expandafter\xdef\csname m\x\endcsname{\noexpand\ensuremath{\noexpand\constantmatrix{\x}}}

	\expandafter\xdef\csname em\x\endcsname{\noexpand\ensuremath{\noexpand\x}}

	\expandafter\xdef\csname rm\x\endcsname{\noexpand\ensuremath{\noexpand\randommatrix{\x}}}

	\expandafter\xdef\csname t\x\endcsname{\noexpand\ensuremath{\noexpand\tensorstyle{\x}}}

	\expandafter\xdef\csname g\x\endcsname{\noexpand\ensuremath{\noexpand\graphstyle{\x}}}

	\ifthenelse{\equal{\x}{E}}{}{
		\expandafter\xdef\csname s\x\endcsname{\noexpand\ensuremath{\noexpand\setstyle{\x}}}
	}

}

\foreach \x in \GreekAlphabet {
	\expandafter\xdef\csname m\x\endcsname{\noexpand\ensuremath{\noexpand\constantmatrixgreek{\csname \x\endcsname}}}

	\expandafter\xdef\csname rm\x\endcsname{\noexpand\ensuremath{\noexpand\randommatrixgreek{\csname \x\endcsname}}}
}

\usepackage[final]{neurips_2023}

\title{Bootstrap Your Own Variance}  %

\newcommand*\samethanks[1][\value{footnote}]{\footnotemark[#1]}

\author{Polina Turishcheva\thanks{Primary contributor. For a detailed breakdown of author contributions see \Cref{sec:contributions}.}\ \ \thanks{Work done during Apple internship.}\ \ $^1$
  \And
  Jason Ramapuram\samethanks[1]\ \ $^2$
  \And
  Sinead Williamson\samethanks[1]\ \ $^2$ \AND
  Dan Busbridge$^2$ \And
  Eeshan Dhekane$^2$ \And
  Russ Webb$^2$ \And
  \vspace{-0.9cm} \\ \phantom{Mystery} \\
  $^1$ University of Göttingen \\
  $^2$ Apple \\ \\
  \footnotesize\texttt{turishcheva@cs.uni-goettingen.de}\\
  \footnotesize\texttt{\{jramapuram, sa\_williamson, dbusbridge, eeshan, rwebb\}@apple.com} \\
\\
}

\begin{document}

\maketitle
\begin{abstract}
Understanding model uncertainty is important for many applications.
We propose Bootstrap Your Own Variance (BYOV), combining
Bootstrap Your Own Latent (BYOL), a negative-free \gls{ssl} algorithm, with Bayes by Backprop (BBB), a Bayesian method for estimating model posteriors.
We find that the learned predictive std of BYOV vs. a supervised BBB model is well captured by a Gaussian distribution, providing preliminary evidence that the learned parameter posterior is useful for label free uncertainty estimation. BYOV improves upon the deterministic BYOL baseline (\textbf{+2.83\%} test ECE, \textbf{+1.03\%} test Brier) and presents better calibration and reliability when tested with various augmentations (eg: \textbf{+2.4\%} test ECE, \textbf{+1.2\%} test Brier for Salt \& Pepper noise).
\end{abstract}

\section{Introduction}
\label{sec:introduction}
Quantifying epistemic uncertainty \citep{hora1996aleatory} is of crucial importance as we increase the use of machine learning models in daily applications \citep{DBLP:journals/corr/abs-2303-08774,DBLP:journals/corr/abs-2204-02311,DBLP:conf/cvpr/RombachBLEO22}.
This task is well suited for Bayesian
machine learning, which replaces point estimates of parameters with a posterior distribution that captures epistemic uncertainty about each parameter's value. While this posterior is typically intractable, we can approximate it using sampling-based methods \citep{metropolis1949monte,neal2011mcmc,izmailov2021bayesian} or \gls{svi} \citep{hoffman2013stochastic}
A Bayesian approach facilitates principled model selection \citep{DBLP:journals/neco/MacKay92, lotfi2022bayesian} and  provides informed decisions that minimize the need for exhaustive hyperparameter searches \citep{snoek2012practical, lotfi2022bayesian}.

Despite its importance, the landscape of uncertainty estimation and calibration within \gls{ssl} remains relatively unexplored, with limited works addressing this critical aspect \citep{hendrycks2019using,bui2022benchmark,DBLP:conf/iclr/GowalHOMK21}, with none taking a Bayesian approach. %
We show that \gls{svi} approaches---specifically, \gls{bbb}---can be used to learn parameter posteriors in SSL, despite the large scale of models and the absence of a likelihood.

The resulting parameter distributions can be used to provide uncertainty  quantification in downstream tasks. It can also give insights about the structure of our model. Modern neural networks are overparameterized \citep{hu2021regularization} and most of the common regularisation or pruning methods today are based only on weight magnitudes \citep{frankle2018lottery}. We show that pruning based on the \gls{snr} ratio of the parameter posterior preserves better performance than magnitude-based pruning in SSL models, extending related findings from the supervised setting \citep{graves2011practical,blundell2015weight}
\subsection{Contributions}
\begin{itemize}[leftmargin=0.75cm]
    \item We propose an algorithm that extends \gls{bbb} to the \gls{ssl} setting. Unlike most Bayesian neural networks that work with small models and datasets \citep[e.g.,][]{blundell2015weight,wang2016natural,wen2018flipout}, we scale \gls{bbb} to Vision Transformers \citep{dosovitskiy2020image}, and train our models on ImageNet-1k \citep{deng2009imagenet}. %
    \item We explore the impact of prior choices on the \gls{byov} posterior, and demonstrate that the resulting uncertainty estimates are  distributionally aligned with outputs from a Bayesian supervised model. %
    \item We compare \gls{snr} pruning \citep{graves2011practical} with magnitude based pruning (without retraining) in the \gls{ssl} setting -- \gls{snr} pruning is up to 12\% better accuracy with a  25\% sparser model. %
\end{itemize}

\section{Background}
\label{sec:background}
\begin{figure}[h]
     \centering
     \begin{subfigure}[b]{0.54\textwidth}
         \centering
         \scalebox{0.72}{\newcommand{\xsep}{1.2}  %
\newcommand{\xysep}{0.2}  %
\newcommand{\ysep}{1.2}  %
\newcommand{\arrowoffset}{0.2} 
\newcommand{\emaoffset}{0.45} 

\newcommand{\arrowlinewidth}{0.6mm}
\newcommand{\newcolor}{AppleRed}

\begin{tikzpicture}[every text node part/.style={align=center}]

\node at (0,0) (x) {$X$};

\node at (\xsep,\ysep+\xysep) (x1t) {$X_1$};
\node at (\xsep,\ysep-\xysep) (x2t) {$X_2$};
\node [draw, rounded corners, rotate=-90] (encodert) at (2*\xsep,\ysep) {Encoder};
\node [draw, rounded corners, rotate=-90] (projectort) at (3*\xsep,\ysep) {Projector};
\node [draw, rounded corners, rotate=-90] (predictort) at (4*\xsep,\ysep) {Predictor};
\node (y1t) at (5*\xsep,\ysep+\xysep) {$y_1^\prime$};
\node (y2t) at (5*\xsep,\ysep-\xysep) {$y_2^\prime$};

\node at (\xsep,-\ysep-\xysep) (x1b) {$X_1$};
\node at (\xsep,-\ysep+\xysep) (x2b) {$X_2$};
\node [draw, rounded corners, rotate=-90] (encoderb) at (2*\xsep,-\ysep) {Encoder};
\node [draw, rounded corners, rotate=-90] (projectorb) at (3*\xsep,-\ysep) {Projector};
\node (y1b) at (5*\xsep,-\ysep-\xysep) {$y_1^\prime$};
\node (y2b) at (5*\xsep,-\ysep+\xysep) {$y_2^\prime$};

\node [rotate=-90] (encoderema) at (2*\xsep+\emaoffset,0) {EMA $\color{\newcolor}(\mu)$};
\node [rotate=-90] (projectorema) at (3*\xsep+\emaoffset,0) {EMA $\color{\newcolor}(\mu)$};

\node (loss) at (6*\xsep,0) {
${\color{\newcolor}\mathbb E_q[}\mathcal L(w,\mathcal D){\color{\newcolor}]-\beta\,\textrm{KL}}$};

\node [draw, rounded corners,color=\newcolor] (sampler) at (3*\xsep,2.5*\ysep) {
$\color{\newcolor}W\sim \mathcal N(\mu,\Sigma=\exp(\theta))$};

\draw[->,-latex,line width=\arrowlinewidth] (x) -- node [above,text width=2cm] {} (x2t);
\draw[->,-latex,line width=\arrowlinewidth] (\xsep+\arrowoffset,\ysep) -- node [above,text width=2cm] {} (encodert);
\draw[->,-latex,line width=\arrowlinewidth] (encodert) -- node [above,text width=2cm] {} (projectort);
\draw[->,-latex,line width=\arrowlinewidth] (projectort) -- node [above,text width=2cm] {} (predictort);
\draw[->,-latex,line width=\arrowlinewidth] (predictort) -- node [above,text width=2cm] {} (5*\xsep-\arrowoffset,\ysep);
\draw[->,-latex,line width=\arrowlinewidth] (y2t) -- node [above,text width=2cm] {} (loss);

\draw[->,-latex,line width=\arrowlinewidth] (x) -- node [above,text width=2cm] {} (x2b);
\draw[->,-latex,line width=\arrowlinewidth] (\xsep+\arrowoffset,-\ysep) -- node [above,text width=2cm] {} (encoderb);
\draw[->,-latex,line width=\arrowlinewidth] (encoderb) -- node [above,text width=2cm] {} (projectorb);
\draw[->,-latex,line width=\arrowlinewidth] (projectorb) -- node [above,text width=2cm] {} (5*\xsep-\arrowoffset,-\ysep);
\draw[->,-latex,line width=\arrowlinewidth] (y2b) -- node [above,text width=2cm] {} (loss);

\draw[->,-latex,dashed,line width=\arrowlinewidth] (encodert) -- node [above,rotate=-90] {} (encoderb);
\draw[->,-latex,dashed,line width=\arrowlinewidth] (projectort) -- node [above,rotate=-90] {} (projectorb);

\draw[->,latex-,line width=\arrowlinewidth,color=\newcolor] (encodert) -- node [above,text width=2cm] {} (encodert |- sampler.south);
\draw[->,latex-,line width=\arrowlinewidth,color=\newcolor] (projectort) -- node [above,text width=2cm] {} (projectort |- sampler.south);
\draw[->,latex-,line width=\arrowlinewidth,color=\newcolor] (predictort) -- node [above,text width=2cm] {} (predictort |- sampler.south);

\draw[->,-latex,line width=\arrowlinewidth,color=\newcolor] (sampler) to[out=0,in=90] ([xshift=1.2cm]loss.north);

\end{tikzpicture}}
         \caption{BYOL and BYOV Architecture.}
         \label{fig:model-schemas}
     \end{subfigure}
     \hfill
     \begin{subfigure}[b]{0.45\textwidth}
         \centering
    \includegraphics[width=.99\textwidth]{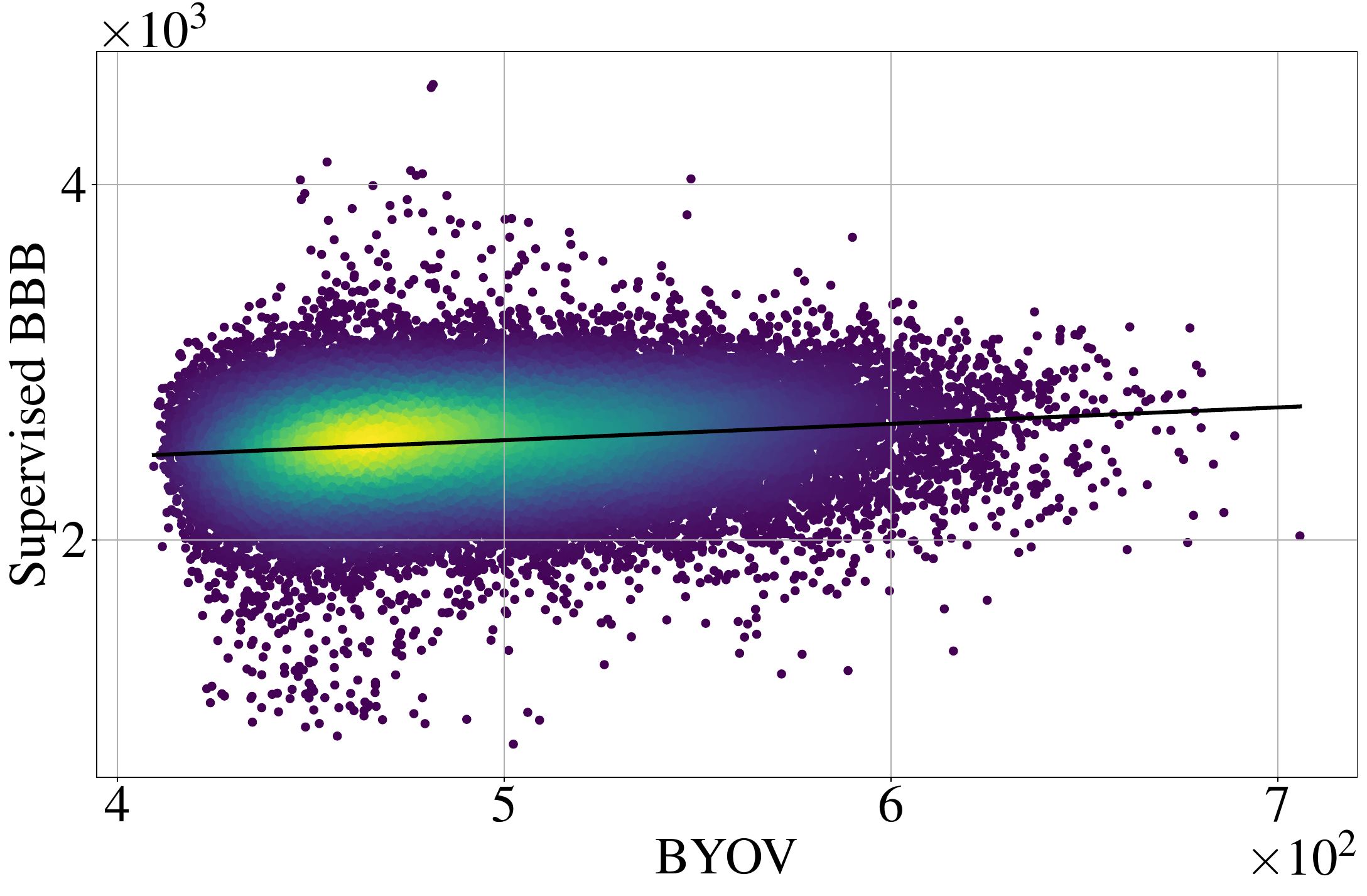}
         \caption{Predictive std of supervised BBB vs BYOV.}
         \label{fig:byol_vs_sup_std}
     \end{subfigure}
    \caption{\textbf{(a)} The standard BYOL architecture is shown in black and modifications required for BYOV are highlighted in red. The BYOV student is parameterized with an Isotropic Gaussian approximate parameter posterior. The teacher is the \gls{ema} of the \gls{map} student parameters.
    \textbf{(b)} Predictive test set standard deviation for supervised BBB versus \gls{byov} overlaid with a Gaussian KDE fit. The predictive std relationship is well captured by a Gaussian distribution, highlighting distributional alignment.
    Models evaluated over the ImageNet1k test set using 1000 MC draws per sample from the approximate parameter posterior, $q(\rvw|\rvtheta)$. Both models are trained with the same $\beta = 0.0 \mapsto 1.0$ schedule.
    }
\end{figure}
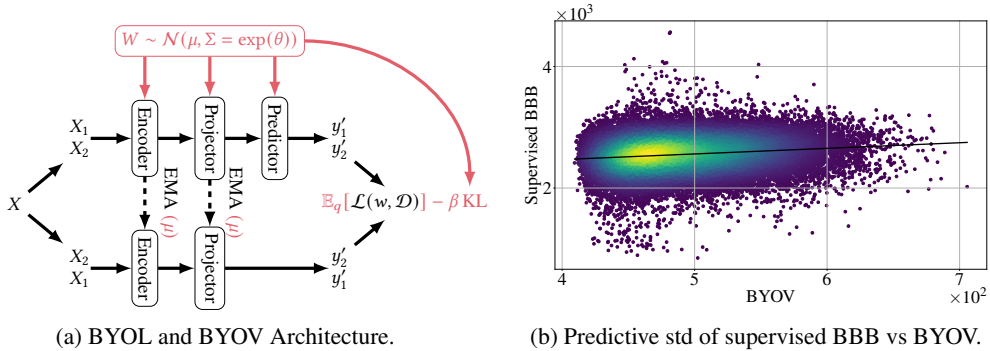
\subsection{Bayes-by-Backprop}
\label{sec:bbb}
Estimation of the parameter posterior, $p(\rvw|\mathcal{D})$, is central to Bayesian
learning.
\gls{bbb} \citep{blundell2015weight} learns the parameters $\rvtheta$ of an approximate posterior, $q(\rvw|\rvtheta)$, by minimizing the KL-divergence against the true  posterior. Since the KL-divergence cannot be evaluated directly, we maximize an alternative objective called the \gls{elbo} \citep{DBLP:journals/neco/DayanHNZ95}%
\begin{equation}\text{ELBO}(\theta;\mathcal{D},\rvw) = \mathbb{E}_{q(\rvw|\theta)}[\log p(\mathcal{D}|\rvw)] - \beta \ \text{KL}\left(q(\rvw|\theta)||p(\rvw)\right), \label{eqn:elbo}\end{equation}
where $\beta>0$ is a Lagrange multiplier \citep{higgins2016beta}. When $\beta=1$, maximizing the \gls{elbo} is equivalent to minimizing $\text{KL}\left(q(\rvw|\theta) || p(\rvw |\mathcal{D})\right)$. In practice, setting $\beta<1$, approximating a \emph{cold posterior}, improves  predictive performance \citep{osawa2019practical,wenzel2020good}.
\subsection{\gls{byol}}
\gls{byol} is a negative-free student-teacher distillation framework that minimizes the cosine similarity between a teacher model and an online student model (\figurename~{\ref{fig:model-schemas}}).
The student comprises three networks: an encoder, a \gls{mlp} projector, and a \gls{mlp} predictor. The teacher model is the exponential moving average of the student encoder and projector. The predictor  introduces an asymmetry between the branches and is a necessary component to prevent collapse \citep{grill2020bootstrap}. \gls{byol} is trained by inferring \emph{two different augmentations of the same image} through both the student and teacher models and minimizing the cosine similarity between the induced representations. After training we can drop the predictor and projector networks and use the student or teacher encoder representations for downstream tasks, such as image classification.
\section{\gls{byov}: A Bayesian SSL method}\label{sec:BYOV}
Here we describe \gls{byov} (Figure~\ref{fig:model-schemas}), which couples \gls{bbb} with \gls{byol}.\footnote{We discuss our choice of \gls{bbb} in Appendix~\ref{sec:discussion} and include more implementation details in Appendix~\ref{app:training_details}.}
\gls{byov} learns a distribution over the parameters of the student model, and uses the student \gls{map} to update the teacher parameters.
BBB typically approximates the posterior distribution over weights, specified in terms of a prior $p(\rvw)$ and a likelihood $p(\mathcal{D}|\rvw)$. However, BYOL does not use a likelihood---our loss is based on the cosine similarity between two representations. Instead, we estimate a \textit{generalized posterior} \citep{bissiri2016general}, $\tilde{p}(\rvw|\mathcal{D}) \propto p(\rvw) \exp\{-\mathcal{L}(\rvw, \mathcal{D})\}$, where $\mathcal{L}(\mathbf{w},\mathcal{D})$ is an arbitrary loss term -- in our case, cosine similarity. We therefore minimize the generalized ELBO \citep{knoblauch2019generalized},\footnote{Note, the term generalized ELBO has been used to describe multiple modifications to the ELBO \citep[e.g.,][]{pmlr-v80-chen18k,domke2018importance}. We specifically refer to the form arising from the ``Rule of Three'' proposed by \citet{knoblauch2019generalized}.} \begin{align}
\text{Generalized ELBO} = \mathbb{E}_{q(\rvw|\rvtheta)}[\mathcal{L}(\rvw, \mathcal{D})] - \beta \ \text{KL}[q(\rvw|\rvtheta) || p(\rvw)] .
\label{eqn:gen_elbo}
\end{align}
In theory, the prior $p(\rvw)$ captures our beliefs about parameter values.
In the context of neural networks, these priors can be hard to define \citep{vladimirova2019understanding,fortuin2021bayesian}.
This has led to many Empirical Bayes methods that learn a data dependent prior \citep{DBLP:conf/aistats/TomczakW18,DBLP:conf/nips/BornscheinMZR17,DBLP:conf/iclr/WuWGL18,DBLP:conf/iclr/RamapuramWK21}.
In this work, we consider three priors: (i) a $\mathcal{N}(0, I)$ used by \citet{blundell2015weight}, (ii) $\mathcal{N}(\mu_T, I)$, using teacher weights as a data informed estimates for the prior means, and (iii) $\mathcal{N}(\mu_T, \Sigma_T)$, where $ \Sigma_T = \text{diag}(\rvtheta_T^2 - \bar{\rvtheta}_T^2)$, using the  variance of the means (for this we keep an \gls{ema} of the second order term $\bar{\rvtheta}_T^2 = \gamma \bar{\rvtheta}^2_T + (1 - \gamma) \rvmu_S^2$.).

\section{Results}
\label{sec:results}
\subsection{Ablations}
\label{sec:ablations}
\begin{figure}
    \centering
    \includegraphics[width=0.95\textwidth]{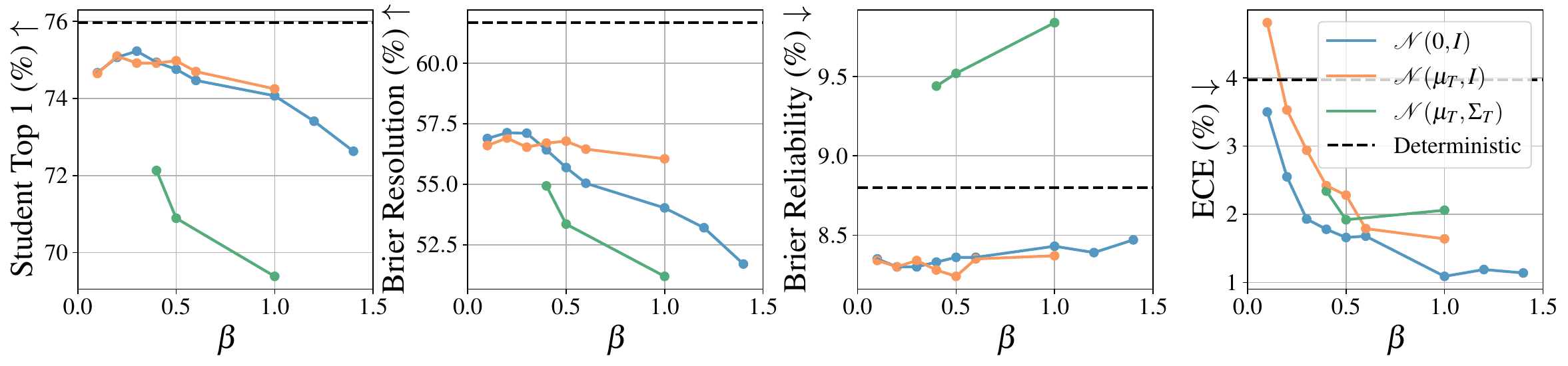}
    \caption{Prior ablation. All metrics here are for in-domain test set using the mean of the student parameter posterior, $\rvmu$, for inference. %
    The best \gls{byov} models outperform the deterministic BYOL model for ECE (\textbf{+2.83\%}) and reliability (\textbf{+1.03\%}), but underperform in top-1 (-0.4\%), top-5 (-0.22\%) and resolution (-0.57\%). }%
    \label{fig:ablation}
\end{figure}
We explore the impact of different prior choices in Figure~\ref{fig:ablation}, which shows accuracy, Expected Calibration Error (ECE) \citep{guo2017calibration} and Brier metrics \citep{gneiting2007strictly} based on the MAP parameters, across a range of values of $\beta$. We find using the posterior mean (obtained by MC sampling) leads to improved results (e.g., 0.34\% higher accuracy with $\mathcal{N}(0, I)$ prior and $\beta=0.3$), but is more expensive to compute. %

The $\mathcal{N}(0, I)$ prior achieves comparable accuracy to a deterministic \gls{byol} model, and improved ECE and reliability. We find little difference between a $\mathcal{N}(0, I)$ prior and a $\mathcal{N}(\rvmu_{T}, I)$ prior, with the former performing slightly better. We hypothesize that this is because the overall performance of the neural network is relatively invariant to constant shifts in the weights. However, we see notably worse performance using $\mathcal{N}(\rvmu_{T}, \bm{\Sigma}_{T})$. This prior actively pulls the student towards the teacher (since $\bm{\Sigma}_{T}$ is typically fairly small), so we hypothesize that this prior does not encourage sufficient difference between teacher and student. In addition, analysis of parameter logs suggest this prior leads to training instabilities, likely because the prior is dynamically varying over training.
\subsection{Exploring the posterior distribution}
\label{sec:layerwise_posterior}
\begin{figure}[H]
    \centering
    \begin{subfigure}[b]{0.48\textwidth}

    \includegraphics[width=\textwidth]{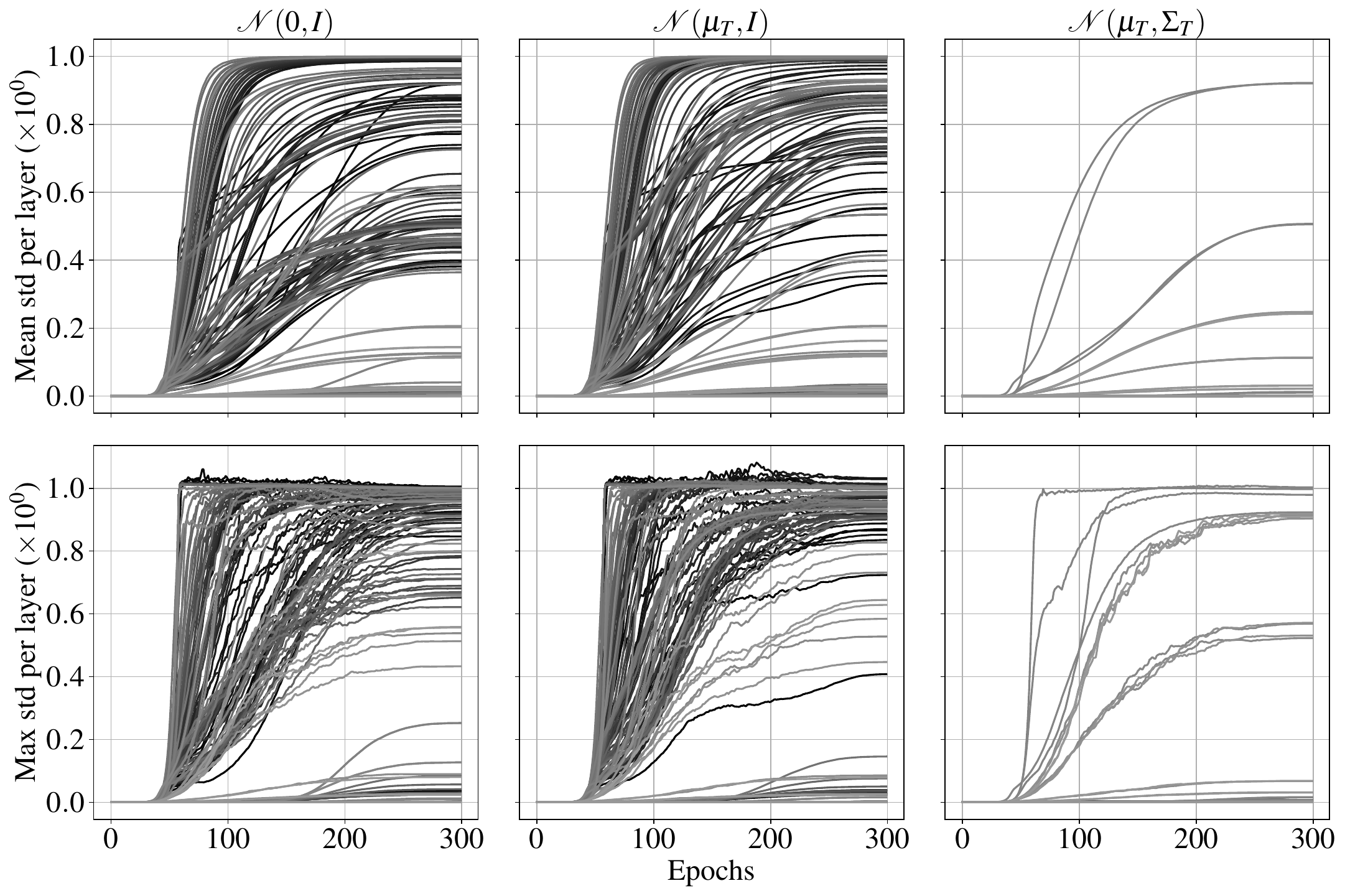}
    \end{subfigure}
    ~
    \begin{subfigure}[b]{0.48\textwidth}
    \includegraphics[width=\textwidth]{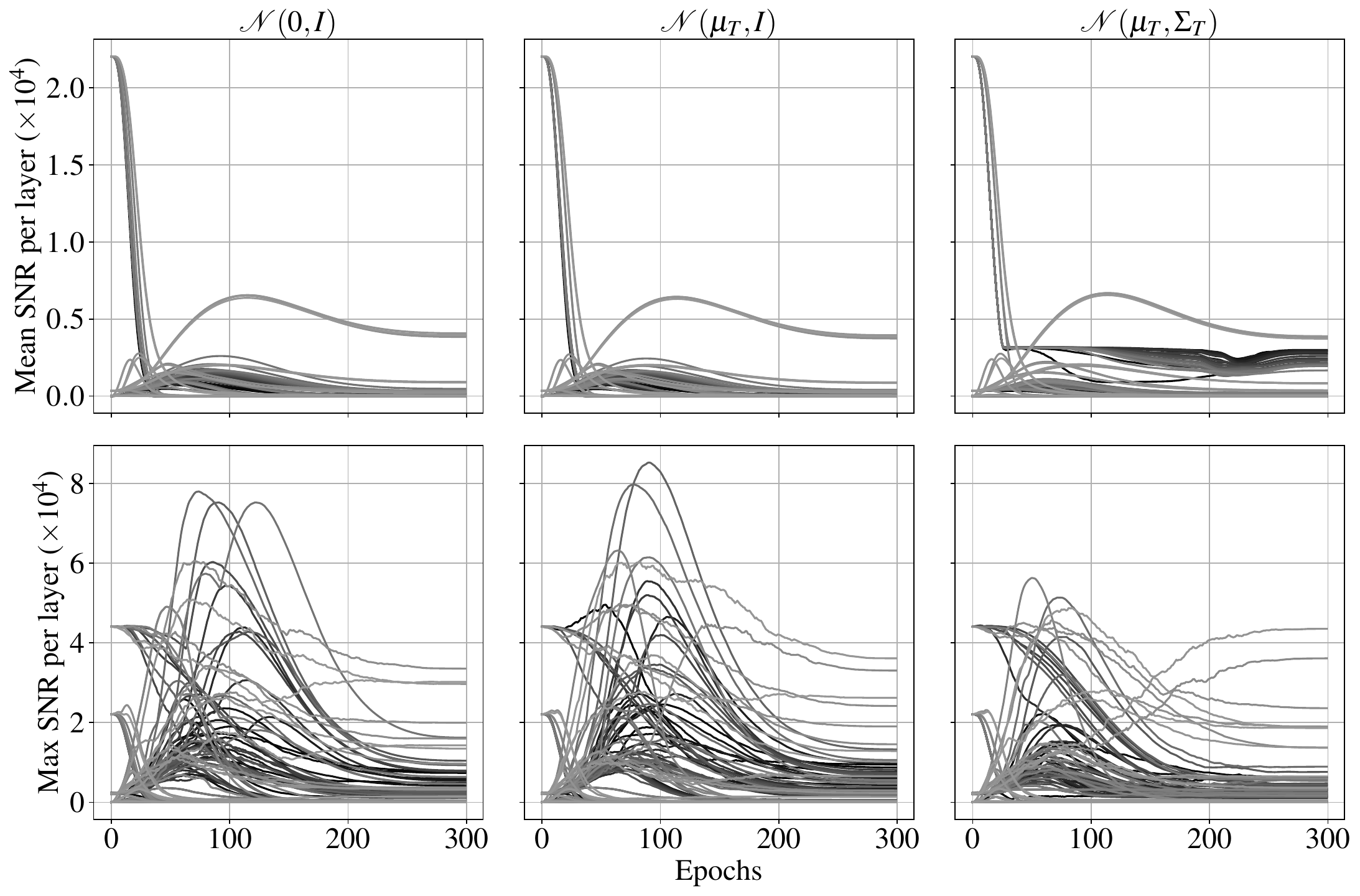}
    \end{subfigure}
    \caption{Prior layerwise $\sigma$ (left) and SNR (right) evolution. All models are trained with $\beta: 0.0 \mapsto 1.0$. Top: layerwise std mean. Bottom: layerwise std max.
    The layerwise $\sigma$ trajectories appear different for three different priors. %
    By contrast, the SNR evolution dynamics (including their maximum values) follow a similar trend.
    }
    \label{fig:different_kld_sigma}
  \end{figure}

Since \gls{bbb} explicitly evaluates the posterior over weights,
we are able to explore the distribution of posterior variance over network layers.
In addition, we are able to analyse the evolution of this uncertainty over training.
In Figure \ref{fig:different_kld_sigma}, we plot the the mean
and maximum value of
the learned standard deviation $\sigma$ and \gls{snr} $|\mu|/\sigma$,
over training for each layer
\footnote{Since \gls{bbb} updates the natural parameters of the parameter posterior at each step of the optimization process, each minibatch will induce a separate weight, which we then aggregate over each dataset epoch.}. We observe that the choice of prior makes a large difference on the learned layer-wise standard deviations.
However, if we look at the posterior \gls{snr} (Figure~\ref{fig:different_kld_sigma}), we see more similarity across priors, particularly between $\mathcal{N}(0, I)$ and $\mathcal{N}(\mu_T, I)$, supporting the idea that performance is relatively invariant to rescaling. In Appendix~\ref{app:snr_vs_beta}, we show that \gls{snr} trajectories remain similar under different choices of $\beta$ (Figure~\ref{fig:different_kld_snr}).%

  \subsection{Does \gls{byov} capture meaningful model uncertainty?}
  \begin{figure}[ht]
    \centering
    \begin{subfigure}[b]{0.37\textwidth}
    \includegraphics[width=\textwidth]{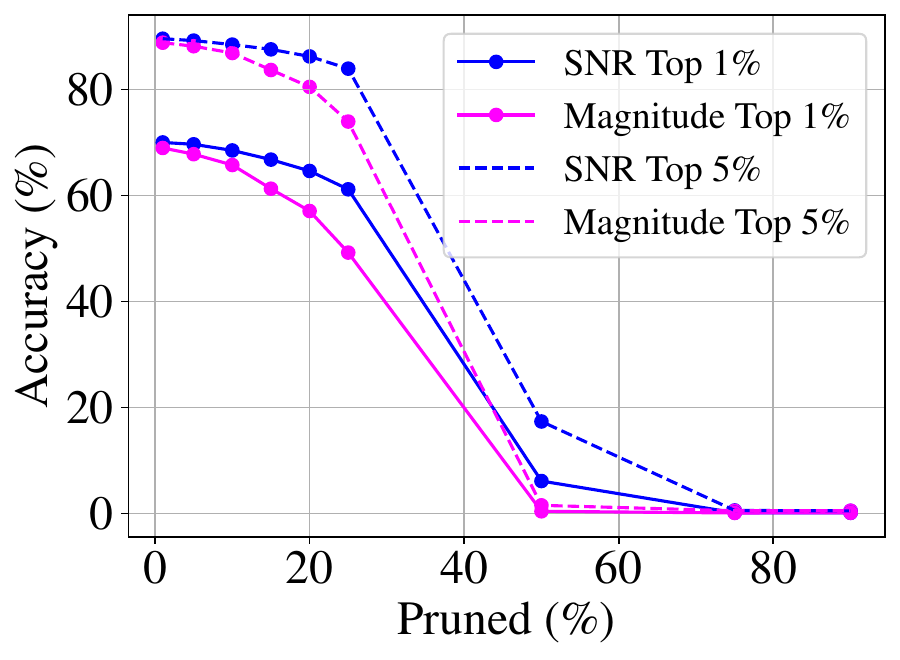}
    \end{subfigure}
    ~
    \begin{subfigure}[b]{0.61\textwidth}
    \includegraphics[width=\textwidth]{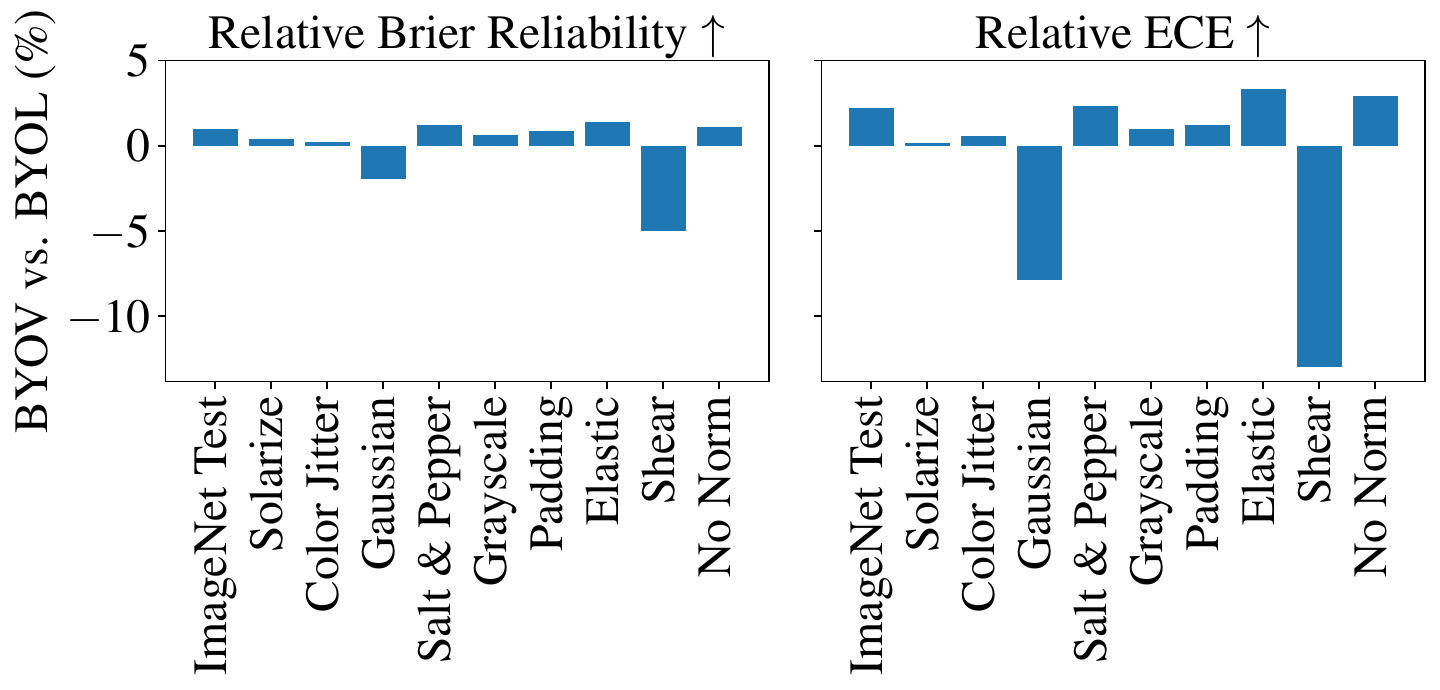}
    \end{subfigure}
    \caption{\textbf{Left}: SNR vs. magnitude based pruning.
    \textbf{Center} : Relative Brier reliability between BYOV and deterministic BYOL.
    \textbf{Right}: Relative ECE between BYOV and deterministic BYOL.
    }
    \label{fig:prunning}
\end{figure}
A natural question that arises is whether the learned posterior distribution captures uncertainties that are relevant to downstream tasks. To assess this, in Figure~\ref{fig:byol_vs_sup_std} we look at the relationship between uncertainty of the \gls{byov} predictive distribution, and uncertainty under a supervised BBB model.  We observe that the relationship between the predictive standard deviation of both models can be suitably captured using a Gaussian, which gives credence to using \gls{byov} as a proxy for the uncertainty of a supervised BBB model.
 We also look at how incorporating BBB impacts prediction quality, by looking at ECE and Brier reliability score. Previous work on supervised models suggests that BBB improves calibration and reliability \citep{ovadia2019can}. In \Cref{fig:prunning} (Center/Right) we also observe improved calibration on the in-distribution data (ImageNet Test). In an out-of-distribution task, we see improved calibration and reliability on many types of augmentation, but notably worse reliability and calibration under shearing and Gaussian augmentations.

\subsection{Pruning}
A high posterior variance indicates that the model lacks confidence in a weight's value. We can use this to prune the network, removing weights where the network lacks confidence. Since network performance can be invariant to weight scale, we follow \citet{blundell2015weight} and use \gls{snr} for pruning, keeping the $x$th percentile per layer. In Figure~\ref{fig:prunning} (Left), we show that this achieves better performance than magnitude-based pruning \citep{frankle2018lottery}, with a sparser model. To simplify our analysis, we do not retrain either model as in \citet{frankle2018lottery}.


\section{Conclusion}
\label{sec:conclusion}
In this work, we introduce \gls{byov}, a method to learn model uncertainty in a label free manner. We explore posterior performance and show that the resulting layerwise \gls{snr} is a good metric for model pruning. We show that posterior variance is correlated with that of supervised models, suggesting the distributions can be used for approximate inference in downstream tasks. %

\bibliography{libraries/autobib,libraries/manualbib}
\bibliographystyle{templates/iclr2021/iclr2021_conference}

\clearpage
\appendix
\appendixpage
\glsresetall

\section{Acknowledgements}
\label{sec:acknowledgements}

We thank 
Arno Blaas, Jonathan Crabbé, Megan Maher, and Amitis Shidani for their helpful feedback and critical discussions throughout the process of writing this paper; Okan Akalin,
Hassan Babaie, 
Denise Hui,
Mubarak Seyed Ibrahim, 
Li Li, 
Cindy Liu, 
Rajat Phull,
Evan Samanas, 
Guillaume Seguin, 
and the wider Apple infrastructure team for assistance with developing and running scalable, fault tolerant code.
Names are in alphabetical order by last name within group.

\section{Discussion -- Why BBB?}
\label{sec:discussion}
We choose \gls{bbb} as our Bayesian estimator because of its theoretical scalability and the findings from  \citet{ovadia2019can}, which highlight the competitive performance of \gls{svi} methods for uncertainty estimation. The authors demonstrate that \gls{bbb} outperforms Expectation Propagation \citep{DBLP:conf/uai/Minka01}, Monte Carlo Dropout \citep{DBLP:conf/icml/GalG16}, and last layer (LL) Bayesian variants \citep{DBLP:conf/iclr/RiquelmeTS18} in terms of \gls{ece}  \citep{guo2017calibration} and Brier score \citep{gneiting2007strictly, brocker2009reliability}.

Although \gls{bbb} presents a concrete algorithm to learn posterior variances, it draws only one posterior weight variate per minibatch. While being an unbiased estimate, this can potentially be high variance.
Towards that end, Flipout \citep{wen2018flipout} aims to mitigate the high variance estimate through weight perturbation. However, since \gls{ssl} methods typically rely on large batch training, we find that this negates the need for such strategies. The authors also confirm this in Appendix E2 \citep{wen2018flipout} where they train with a batch size of 8192 which is equivalent in our setting.

\section{Training details}\label{app:training_details}
\paragraph{Modifications to \gls{byol}}
To keep consistency with the baseline \gls{ssl} method we closely follow the model architecture and hyper-parameters defined for \gls{byol} \citep{richemond2020byol} with alterations made to support Vision Transformers \citep{dosovitskiy2020image,busbridge2023scale}. However, naively applying \gls{bbb} to \gls{byol} does not work out-of-the-box and required  following changes: %
\begin{itemize}

    \item \textbf{Removal of weight decay}: \gls{byol} default recipe includes weight decay.
    We remove weight decay as it interferes with the learning dynamics when coupling \gls{bbb} with \gls{byol}. The KL loss term already introduces a regularisation effect and
    explicitly pulls towards a prior. %

    \item \textbf{KL annealing}: a commonly applied practice in latent variable stochastic inference is to use a schedule for the KL divergence \citep{DBLP:conf/nips/SonderbyRMSW16}. We find that this also helps to improve the downstream tasks performance in \gls{byov} paradigm.

\end{itemize}

In addition, we made a number of changes to the \gls{bbb} algorithm, in order to encourage stability:
\begin{itemize}
    \item \textbf{Initialization of $\rvsigma^2$}: When learning $\rvsigma^2$ as a free parameter a non-negativity constraint needs to be enforced, we use exponential function for this whereas \cite{blundell2015weight} uses Softplus. Log-variances are initialized with -10. Means are initialised using trunc\_normal(std=0.02) \citep{randaug_touvron2021training}.
\textbf{Other details}: To keep consistency with the baseline \gls{ssl} method we closely follow the model architecture and hyper-parameters defined for \gls{byol} \citep{richemond2020byol} with alterations made to support Vision Transformers \citep{dosovitskiy2020image,busbridge2023scale}.
\item \textbf{Scheduling $\beta$}. Previous work has indicated that annealing the $\beta$ weight applied to the KL term in the ELBO from zero to the desired value yields improved performance over using a fixed value of $\beta$. We found this to be the case in practice.
\end{itemize}

\section{MC number of samples estimate}
\label{sec:mc}
\begin{figure}
    \centering
    \includegraphics[width=\textwidth]{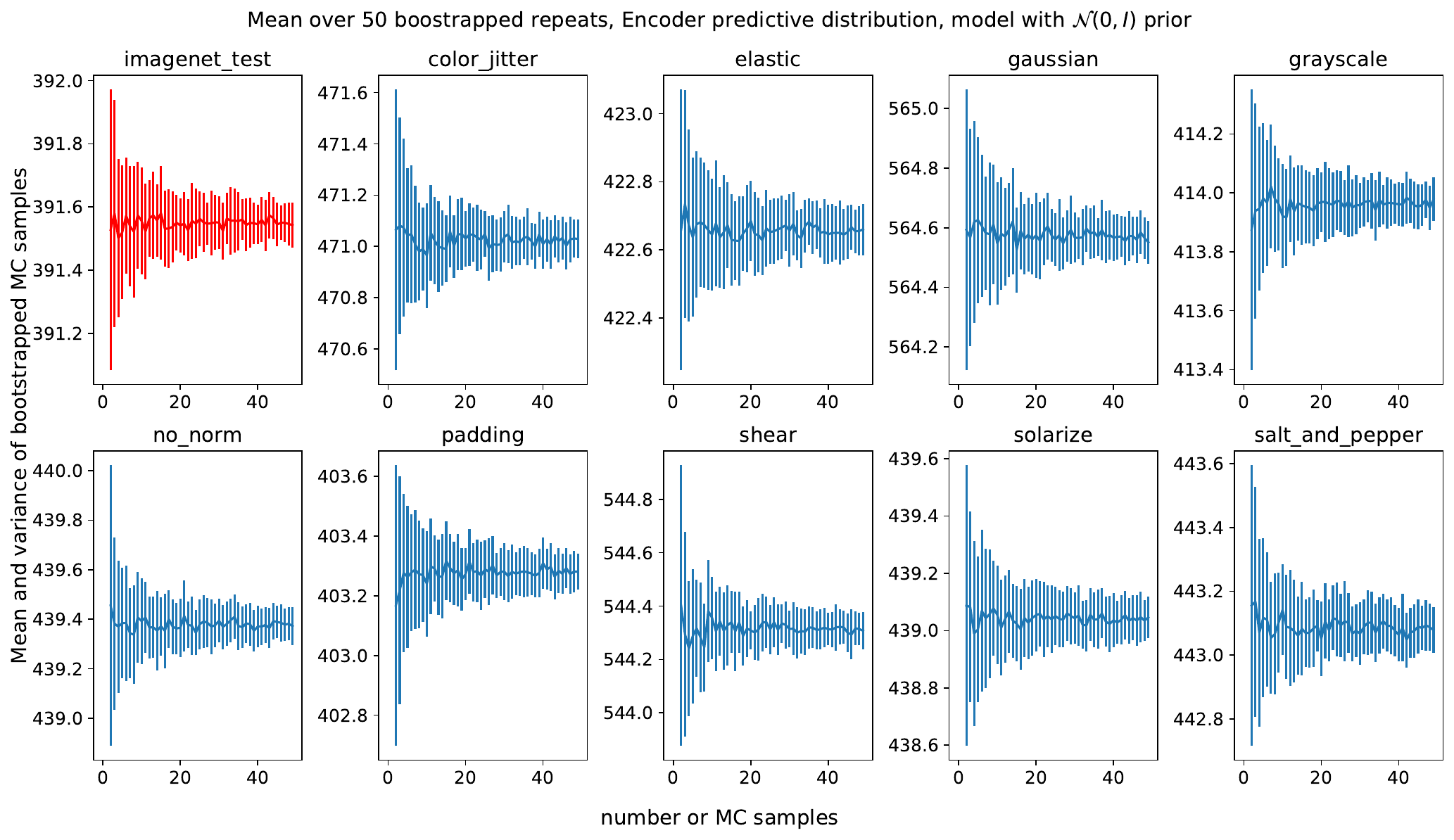}
    \caption{Mean and standard deviation of predictive distribution under different dataset augmentations. Model was trained with a scheduled $\beta: 0.0 \mapsto 1.0$. The standard deviation of the expectation converges to 0 and the expectation appears stable. This is consistent across models. }
    \label{fig:ms-sampling}
\end{figure}

To evaluate the posterior predictive distribution we infer each input sample, $x \sim p(\rvx)$, using $K$ draws from the parameter posterior, $\{\rvw_i\}_{i=1}^K \sim q(\rvw|\rvtheta)$. Previous work uses thirty MC draws \citep{maddox2019simple,lotfi2022bayesian,daxberger2021laplace}, but does not justify the validity of this decision. We ablate this in Figure \ref{fig:ms-sampling} using the entire test set of ImageNet1k (50,000 samples). We sample the posterior from one to fifty times per sample ($\times 50$ bootstrap) and evaluate the predictive mean and standard deviation. To provide a tighter estimate, we use 1000 MC draws in Figure \ref{fig:model-schemas}(b).

\section{Exploring the posterior distribution: Additional results}\label{app:snr_vs_beta}
\begin{figure}[h]
    \centering
    \includegraphics[width=\textwidth]{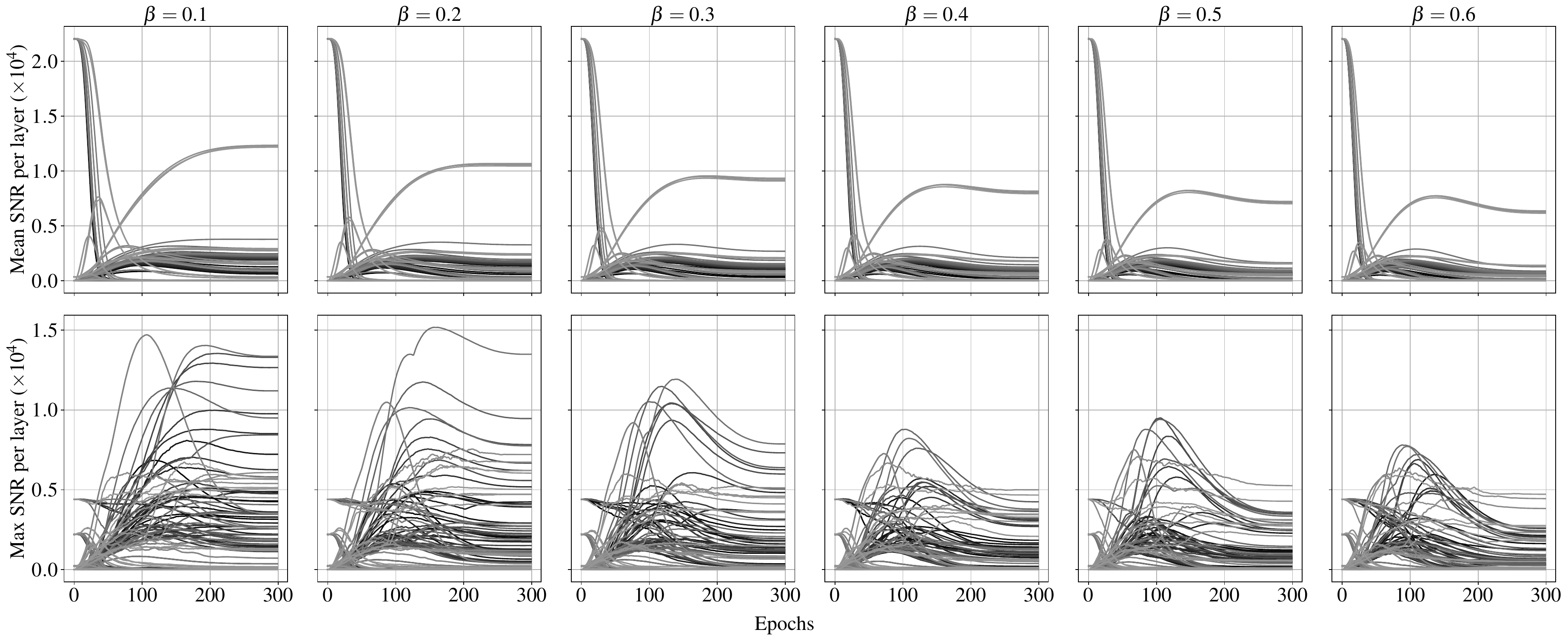}
    \caption{
    SNR ( $|\mu | / \sigma$ ) with different KL-$\beta$ schedules. All schedules start from 0.0 and follow a single cycle cosine to end at the $\beta$ described in the title. All models present similar trajectories over training, but vary slightly for their maximum (bottom row).
    The values and patterns are mostly identical between all three models. We can see a trend for the growing bundle of lines that achieve smaller values along with increasing $\beta$. These lines correspond to the projector and predictor.
    }
    \label{fig:different_kld_snr}
\end{figure}
In Section~\ref{sec:layerwise_posterior}, we looked at how SNR varied across different priors, showing the layerwise SNR is relatively invariant to the choice of prior. In Figure~\ref{fig:different_kld_snr}, we repeat this analysis using varying values of the KL weight $\beta$ (with a $\mathcal{N}(0, 1)$ prior). We find that $\beta$ has a fairly small impact on the SNR values.

\section{Increasing Gaussian noise}
\begin{figure}[ht]
    \centering
    \includegraphics[width=\textwidth]{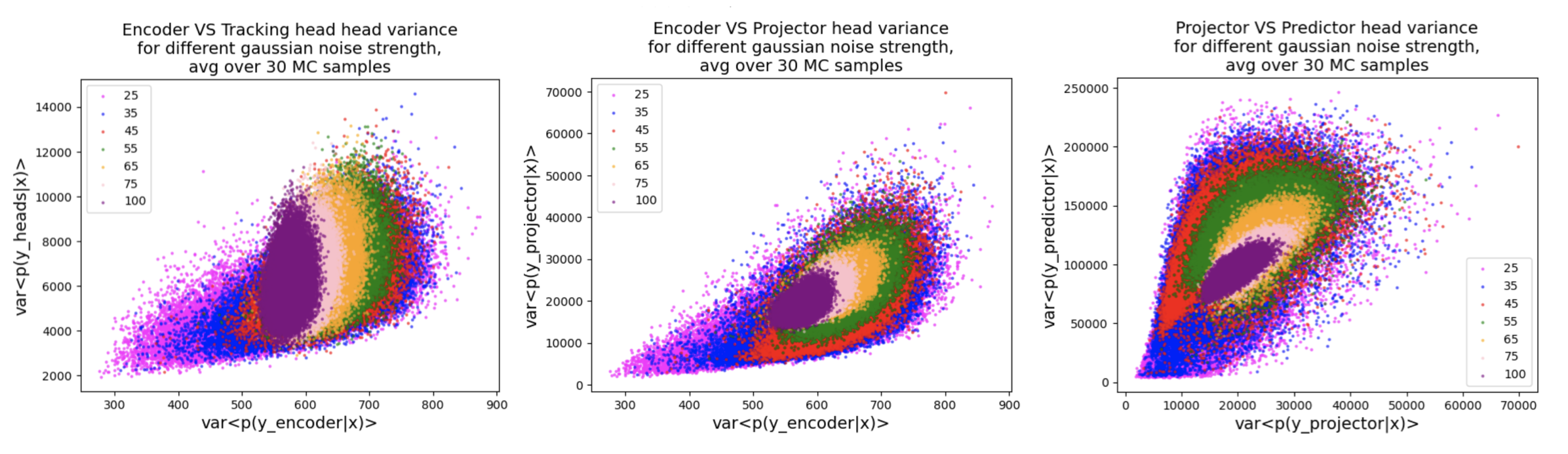}
    \caption{Variance of latent variables from different model parts. The stonger gaussian noise is, the less meaningful features can be extracted from the encoder, hence, the variances of predictive distribution become more concentrated. Each point is based on  $M=50$ weights MC sample (App.\ref{sec:mc}).}
    \label{fig:variances_from_dif_parts}
\end{figure}

Figure~\ref{fig:variances_from_dif_parts} shows posterior predictive variances obtained at three points in the model: after the encoder layer; after the projector, and after the predictor heads (see Figure~\ref{fig:model-schemas}).
Each point represents an image augmented with Gaussian noise; the color of each point represents the strength of the augmentation. We see positive correlation between uncertainty at each location. Moreover, as the amount of noise increases, the average variance increases, as we would expect. Meanwhile, the variation in the variances decreases, as images become increasingly close to pure Gaussian noise, hence, the variance become more concentrated.

\section{Should the whole network be Bayesian?}
\label{sec:net_be_bayesian}
Previous work has also questioned whether an entire network needs to be Bayesian \citep{sharma2023bayesian, gast2018lightweight, ovadia2019can}.
A common reason for a partial Bayesian network is due to the incurred memory overhead used to learn the natural parameters of the distribution. While an Isotropic Gaussian \gls{bbb} doubles parameter counts, the effective memory footprint of the model does not grow proportionally. Typically, the majority of accelerator memory is dominated by activation gradients and not model parameters \citep{DBLP:journals/corr/ChenXZG16}. After reparameterization \citep{DBLP:journals/jmlr/MohamedRFM20}, the effective activation size is equivalent to the \gls{mle} case for every layer, thus incurring only a minimal practical overhead for being Bayesian.

In Table \ref{tab:ablation}, we retrain a subset of \gls{byov} models by keeping part of the network Bayesian. In particular, we explore using a point estimate for all LayerNorm layers and the convolutional patcher used in vision transformers. In contrast to previous findings we observe that using a fully Bayesian model (Fig. \ref{fig:ablation}) presents the best performance amongst the class of \gls{bbb} models.

\begin{table}[ht]
    \centering
    \begin{tabular}{ccccccccc}
         BBB & Prior & $\beta_{start}$ & $\beta_{end}$ & top-1 $\uparrow$ & top-5 $\uparrow$ \\
        conversion & Type &  &  &  (student) & (student) \\
        \hline

        No Conv & $\mathcal{N}(0, I)$  & 0.0 &  1.0 & 73.68 & 91.43 \\
        BBB & $\mathcal{N}(\mu_{T}, I)$  &  &   & \textbf{73.97} & 91.52 \\
        & $\mathcal{N}(\rvmu_{T}, \bm{\Sigma}_{T})$  &  &   & 69.73 & 88.74 \\
        \hline
        Linear & $\mathcal{N}(0, I)$  & 0.0 &  1.0 & 73.62 & 91.12 \\
        Only BBB & $\mathcal{N}(\rvmu_{T}, I)$  &  &  & 74.23 & 91.48 \\

        & $\mathcal{N}(0, I)$  & 1.0 &  1.0 & 72.60 & 90.92 \\
        & $\mathcal{N}(\rvmu_{T}, I)$  &  &  & 72.69 & 91.03 \\
        \specialrule{.2em}{.1em}{.1em}
        No BBB (baseline) & & &  & 75.97 & 92.40\\
        \hline
    \end{tabular}
    \caption{All metrics are computed on the in-domain test set and uses the posterior mean, $\rvmu$, for inference.}
    \label{tab:ablation}
\end{table}
\FloatBarrier
\section{Contributions}
\label{sec:contributions}
All authors contributed to writing this paper, designing the experiments, discussing results at each stage of the project.

\paragraph{Preliminary work} Formulation of \gls{bbb} coupled with \gls{ssl} developed by Polina Turishcheva, Jason Ramapuram and Sinead Williamson. Idea refined in discussions with Dan Busbridge, Eeshan Dhekane and Russ Webb.

\paragraph{Generalized ELBO formulation} Relationship of ELBO to the generalized posterior and related formulations developed by Sinead Williamson (\Cref{sec:bbb}, \Cref{sec:BYOV}).

\paragraph{Pruning} Experiments written by Polina Turishcheva in discussions with Russ Webb (\Cref{fig:prunning})-Left.

\paragraph{Briar reliability and ECE analysis} Conducted by Polina Turishcheva (\Cref{fig:prunning} - Center, Right and \Cref{fig:variances_from_dif_parts}).

\paragraph{Layerwise variance and SNR exploration} Conducted by Polina Turishcheva in discussions with Sinead Williamson and Jason Ramapuram (\Cref{fig:different_kld_sigma}, \Cref{fig:different_kld_snr}).

\paragraph{Monte Carlo Variance Estimates} Preliminary explorations into Monte Carlo repeats (\Cref{fig:ms-sampling}) and estimating predictive distribution done by Polina Turishcheva in discussions with Sinead Williamson and suggestions from Dan Busbridge and Russ Webb  (\Cref{fig:variances_from_dif_parts}). Large sample Monte-Carlo estimate experiment (\Cref{fig:byol_vs_sup_std}) and improvements noted in \Cref{sec:ablations} for top-1 done by Jason Ramapuram.

\paragraph{Data dependent priors} Discussions between Jason Ramapuram, Polina Turishcheva, Sinead Williamson and Dan Busbridge led to exploring various priors (\Cref{fig:ablation}). Code written by Jason Ramapuram and validated by Eeshan Dhekane.

\paragraph{Should the whole network be Bayesian?} Explored by Jason Ramapuram in discussions with Polina Turishcheva and Sinead Williamson (\Cref{sec:net_be_bayesian}).

\paragraph{Implementation details} Code for baseline BYOL ViT written by Jason Ramapuram. \gls{byov} implementation written by Polina Turishcheva and Jason Ramapuram. Reviewed by Eeshan Dhekane. Tikz wizardry done by Dan Busbridge.

\end{document}